\documentclass[runningheads]{llncs}

\usepackage{eccv}

\usepackage{eccvabbrv}

\usepackage{graphicx}
\usepackage{booktabs}
\usepackage{multirow}
\usepackage[misc]{ifsym}
\usepackage[accsupp]{axessibility}  

\usepackage{hyperref}

\newcommand{\corrauth}{\textsuperscript{\textrm{\Letter}}}

\usepackage{orcidlink}

\begin{document}

\title{FuDU: A Fuzzy Dual-dimensional Uncertainty Framework for Streaming Active Learning in Industrial Defect Detection}


\titlerunning{FuDU: Fuzzy Dual-Dimensional Uncertainty}

\author{Zhaoyang Wang\inst{1}\orcidlink{0000-0002-4944-3924} \and
Haiyong Chen\inst{1}\corrauth\orcidlink{0000-0002-5262-4208} \and
Binyi Su\inst{1}\orcidlink{0000-0002-8024-347X} \and
Xinwei Lyu\inst{1}\orcidlink{0009-0000-0976-4590}}

\authorrunning{Z.~Wang et al.}

\institute{
1. School of Artificial Intelligence, Hebei University of Technology, Tianjin, China\\
\email{haiyong.chen@hebut.edu.cn}
}

\maketitle

\begin{abstract}
Ensuring the reliability of deep learning models in real-time industrial defect detection is critical for high-stakes quality inspection. To mine uncertain samples within continuous industrial media streams, thereby enhancing the reliability of the detection system, this paper proposes a streaming active learning method based on the Fuzzy Dual-dimensional Uncertainty (FuDU) framework. Specifically, we first design a Prototype-based Global Uncertainty Quantification (PGUQ) module on the backbone to evaluate image-level uncertainty via normal/defective feature prototypes. A Dual-entropy defect Uncertainty Evaluator (DeUE) is then integrated into the detection head to quantify box-level uncertainty. Finally, by modeling uncertainty as systematic error, we propose a fuzzy dual-dimensional uncertainty-aware strategy that leverages fuzzy inference to fuse dual-dimensional uncertainties, enabling expert knowledge-driven adaptive sampling decisions. Comprehensive experiments demonstrate that FuDU is efficient and flexible, making it well-suited for challenging industrial inspection tasks such as the detection of nuclear fuel rod defects. Our code is publicly available at: \url{https://github.com/wangzhaoyang-508/FuDU}.
  \keywords{Industrial defect detection \and Streaming active learning \and Uncertainty quantification}
\end{abstract}

\section{Introduction}
\label{sec:intro}
\begin{figure}[!tb]
\centering
\includegraphics[width=0.6\linewidth]{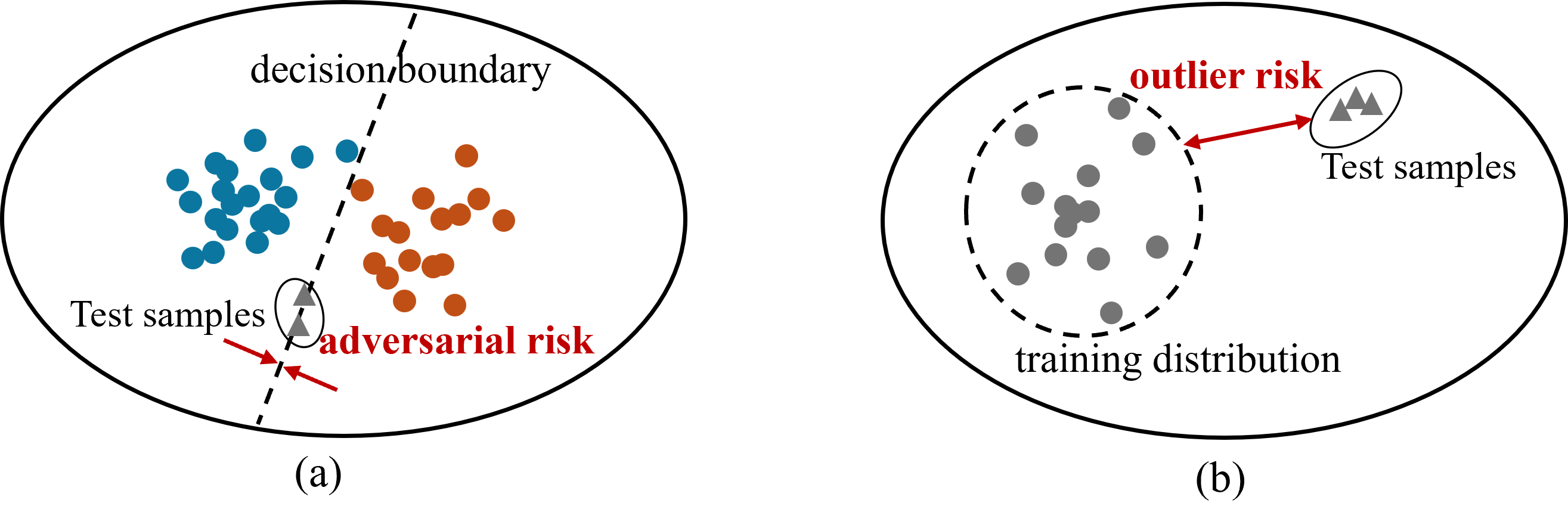}
\caption{Illustration of adversarial and outlier risks.}
\label{fig:Image1}
\end{figure}

While deep learning-based visual models enable efficient defect detection, their efficacy hinges on the availability of large-scale and well-annotated datasets \cite{fu2024cross, yang2024defect}. In many industrial scenarios, defect samples are inherently scarce and typically collected in static laboratory settings. This exposes the deployed models to two critical risks: The first is the \textbf{adversarial risk} stemming from samples near the decision boundary, as shown in \cref{fig:Image1} (a). Specifically, features are close to the decision boundary, making it difficult to determine whether a sample is defective or to precisely what category of defects it belongs to. The second involves the \textbf{outlier risk} arising from samples outside the training distribution, as illustrated in \cref{fig:Image1} (b); here, novel defects or domain shifts cause image features to become outliers that deviate from the original distribution \cite{10540405, zhang2020towards}. To mitigate these risks, researchers have explored various learning paradigms, including few-shot learning \cite{shi2024few}, incremental learning \cite{tang2024incremental}, anomaly detection \cite{chen2024unified}, domain adaptation \cite{cao2024adaclip}, and language-guided foundation models for open-set detection and zero-/few-shot anomaly recognition \cite{xu2025anomalyov,wang2026maugpt}. However, a more direct approach involves actively selecting high-risk uncertain samples during inference for expert annotation. This process iteratively optimizes the decision boundary and enhances the model's domain generalization capability—a paradigm known as Active Learning (AL) \cite{10537213, 11092693}.

\begin{figure*}[!tb]
\centering
\includegraphics[width=\linewidth]{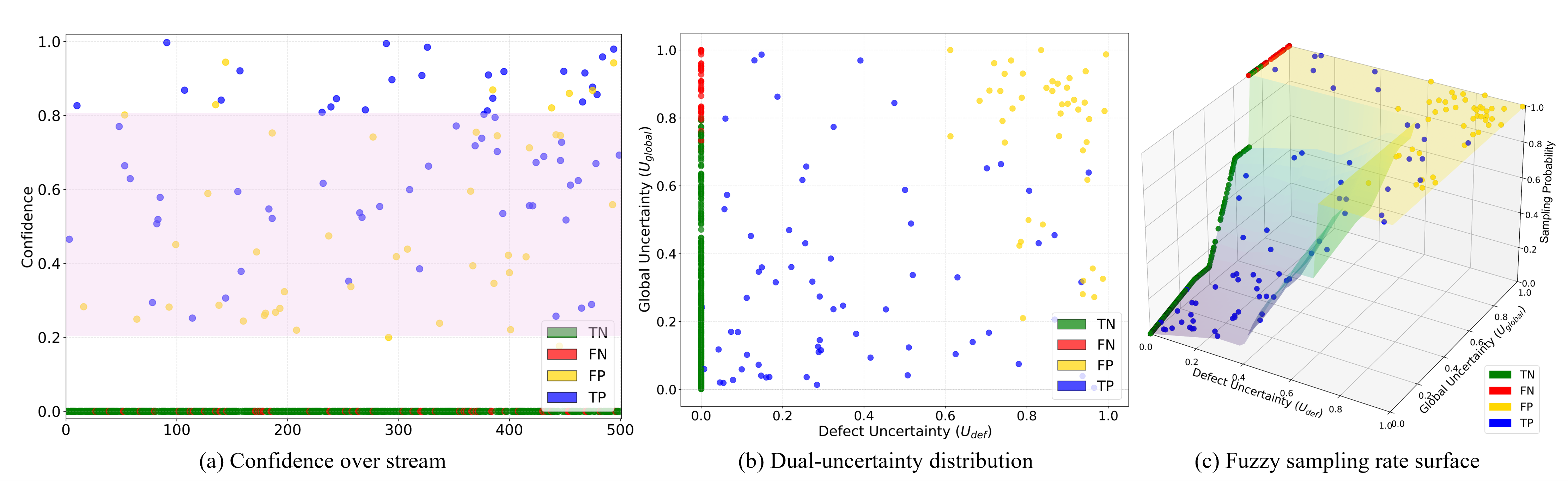}
\caption{Comparison of different uncertainty perspectives and sampling strategies on 500 fuel rod images. (a) Image distribution and sampling strategy based on confidence; (b) The image distribution in the dual-uncertainty space; (c) Sampling probability surface based on our FuDU strategy.}
\label{fig:Image2}
\end{figure*}

The key to AL lies in uncertainty quantification and sampling strategies. In the context of industrial defect detection, while the most straightforward approach involves selecting images containing low-confidence defects as uncertain samples, false negatives compromise the value of model iteration. \cref{fig:Image2} (a) shows the prediction results of the RT-DETR detector \cite{RTDETR} on 500 nuclear fuel rod inspection images, where the y-axis represents confidence scores and the pink region indicates the confidence-based sampling range. As can be seen, although confidence-threshold AL can collect numerous false-positive samples (yellow), it fails to identify false negatives (red). Furthermore, existing pool-based AL methods, despite their effectiveness, fail to satisfy real-time constraints \cite{streamsAL}. 

To overcome these limitations and enable reliable AL within nuclear fuel rod defect detection, we propose a Fuzzy dual-dimensional uncertainty (FuDU) framework. As shown in \cref{fig:Image3}, FuDU incorporates a Prototype-based Global Uncertainty Quantification (PGUQ) module into the detector's backbone. This module assesses global uncertainty $U_g$ as outlier risk by leveraging dynamic feature prototypes of normal and defect samples (i.e., samples exhibiting large distances from both prototypes are identified as outliers). Additionally, the detection head integrates a Dual-entropy Uncertainty Evaluator (DeUE) that computes defect uncertainty $U_d$ as adversarial risk using box localization and classification entropy (i.e., higher entropy values correspond to elevated adversarial risk). As shown in \cref{fig:Image2} (b), $U_g$ and $U_d$ better characterize the model’s uncertainty, facilitating the identification of FN and FP samples. Finally, we introduce a fuzzy dual-dimensional uncertainty evaluation strategy that maps $U_g$ and $U_d$ to linguistic variables and employs fuzzy inference to determine the sampling decision, as shown in \cref{fig:Image2} (c).

Extensive experiments verify FuDU’s effectiveness and scenario adaptability. In our nuclear fuel rod defect dataset, the recall reaches 99.3\% when combined with uncertainty-based sampling. Compared to manual screening, FuDU reduces the annotation cost to less than one-sixth. Furthermore, it outperforms the confidence-based baseline method by 7.3\% in mAP. Its cross-scenario generalization is further validated on the public ELES dataset \cite{SSN}. Notably, the linguistic variables in FuDU allow quality inspection experts to interpret, validate, and calibrate the sampling strategy, effectively integrating domain expertise into the model. Our key contributions are summarized as follows:

\begin{itemize}
\item To improve the reliability of industrial visual inspection systems, we propose FuDU—a fuzzy dual-dimensional uncertainty-driven streaming active learning framework that integrates domain expertise to jointly evaluate defect and global uncertainty for efficient sampling and model updating. To our knowledge, FuDU is the first work to incorporate fuzzy control theory into streaming active learning for industrial defect detection.

\item We design two complementary uncertainty quantification modules: PGUQ and DeUE. PGUQ leverages feature prototypes to identify potential FNs and domain shifts, while DeUE captures borderline or novel defects by quantifying box-level uncertainty via dual entropy and confidence.

\item Extensive experiments demonstrate that FuDU outperforms state-of-the-art AL methods, ensuring the detector’s reliability and iterative efficiency, making it well-suited for high-stakes industrial visual inspection scenarios.
\end{itemize}

\section{Related Work}
\subsection{Industrial Defect Detection}
In general, deploying general-purpose object detection models for industrial defect detection faces challenges such as severe class imbalance and limited operational scenarios, which compromise model reliability. Although some works have proposed methods such as few-shot learning \cite{wang2023meta,su2025few,9395571}, domain adaptation\cite{10819359,10938257,Xinwei}, incremental learning \cite{wang2024multi,ge2025class}, or the use of pre-trained models \cite{cpdd,10922728}, it remains challenging to simultaneously address the two risks illustrated in \cref{fig:Image1}. For high-risk scenarios involving large streams of unlabeled data, employing AL to iteratively enhance model reliability from a data-centric perspective is a more practical approach. Although one-class anomaly detection (AD) \cite{shi2025industrial,aqeel2025towards} quantifies anomalies via normal prototypes and feature distances, its lack of defect categorization and threshold sensitivity limit applicability. Nonetheless, this paradigm inspires our PGUQ for outlier risk assessment.

\subsection{Active Learning for Object Detection}
Active Learning (AL) aims to achieve maximal model performance with minimal annotation cost by iteratively selecting the most informative samples for labeling \cite{10537213}. In object detection, existing AL methods mainly rely on uncertainty- and diversity-based sampling, using techniques such as MC dropout \cite{feng2019deep,gal2017deep,choi2021active}, query-by-committee \cite{10812842,vo2022active}, evidential learning \cite{park2023active}, pseudo-loss estimation and instance-level constraints \cite{yang2025active}, feature clustering \cite{PPAL,fast}, and coreset selection \cite{CSOD,OFDS}. However, most of these methods are tailored to pool-based settings, requiring offline processing of unlabeled samples. In contrast, industrial inspection requires AL under streaming data, real-time constraints, and severe class imbalance, making pool-based methods difficult to deploy directly \cite{streamsAL}.

\subsection{Fuzzy Control and Uncertainty-Aware Decision Making}
Fuzzy control theory~\cite{zadeh1965fuzzy} enables reasoning under uncertainty via linguistic variables and IF-THEN rules. Variants like Mamdani~\cite{mamdani1975experiment} and Takagi-Sugeno~\cite{takagi1985fuzzy} are widely adopted for their interpretability and robustness~\cite{precup2006stability}. In active learning, fuzzy logic facilitates the fusion of heterogeneous uncertainty metrics, enabling dynamic threshold adjustment in streaming scenarios~\cite{7820039}. Crucially, fuzzy systems naturally integrate expert knowledge~\cite{gu2023autonomous}, which is essential for safety-critical inspection where black-box decisions are unacceptable~\cite{rudin2019stop}. In industrial visual inspection, fuzzy rules enhance noise tolerance and trustworthiness compared to hard-threshold methods~\cite{kim2025fuzzy}. Motivated by these advantages, we design a domain-expert-calibrated fuzzy inference module to jointly reason over global and local uncertainties for robust streaming AL.

\section{Methodology}
\subsection{Overview}
\begin{figure*}[!tb]
\centering
\includegraphics[width=\linewidth]{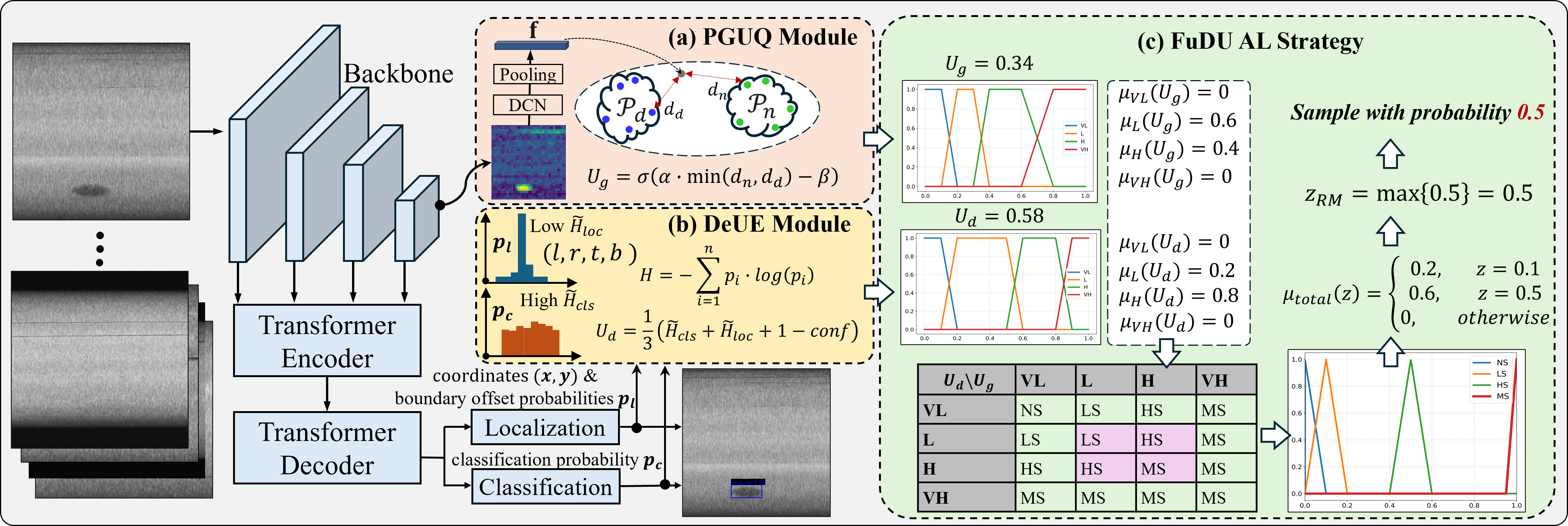}
\caption{Overview of the proposed FuDU framework.}
\label{fig:Image3}
\end{figure*}

As shown in \cref{fig:Image3}, the FuDU framework comprises a backbone, a transformer encoder-decoder, and a detection head. During detection, for each image in the data stream, FuDU yields not only inference results but also an image-level uncertainty score $U_g$ based on PGUQ and a defect-level uncertainty score $U_d$ based on DeUE. FuDU determines the sampling probability for the current image based on a pre-defined fuzzy inference system and the fusion of $U_g$ and $U_d$. After each detection batch, the collected uncertain images are annotated and utilized to train and update both the detector and the prototype library.

\subsection{Prototype-based Global Uncertainty Quantification}
Variations in the production line induce domain shift, manifested as changes in the distribution of image features, leading to increased global uncertainty of the model. To actively perceive and quantify this, we propose a Prototype-based Uncertainty Quantification module (PGUQ). As illustrated in \cref{fig:Image3} (a), PGUQ extracts a feature prototype $\mathbf{f}$ from the feature maps via deformable convolution and global average pooling. The core hypothesis of PGUQ relies on density-based uncertainty estimation: samples located far from high-density regions (represented by prototypes) are considered to have higher uncertainty. We compare $\mathbf{f}$ against normal/defective prototypes to obtain the global uncertainty $U_g$, calculated as follows:

\begin{equation}
U_{g} = \sigma\!\left( \alpha \cdot d_{\min} - \beta \right), \quad
d_{\min} = \min(d_n, d_d)
\end{equation}

\noindent where the distances to the nearest normal and defective prototypes are defined as:
\begin{equation}
d_n = \min_{\mathbf{p} \in \mathcal{P}_n} \|\mathbf{f} - \mathbf{p}\|_2, \quad
d_d = \min_{\mathbf{p} \in \mathcal{P}_d} \|\mathbf{f} - \mathbf{p}\|_2
\end{equation}

\noindent Here, $\mathbf{f}$ denotes the feature prototype of the current image, while $\mathcal{P}_n$ and $\mathcal{P}_d$ represent the sets of normal and defect prototypes, respectively. $d_{\min}$ measures the proximity of the current sample to the closest known pattern. We employ learnable parameters $\alpha$ and $\beta$  along with a sigmoid function $\sigma(\cdot)$ to normalize $U_g$ to $[0, 1]$. A value closer to 1 indicates the sample lies in a low-density region, implying greater model uncertainty.

The prototype sets $\mathcal{P}_n$ and $\mathcal{P}_d$ are initialized by applying K-means clustering to features extracted from the initial training set. This provides a prior distribution of the feature manifold. Crucially, to handle domain shift during AL, we treat prototypes as learnable parameters optimized end-to-end via backpropagation. Specifically, gradients flowing through the distance metrics during training enable prototypes to shift towards the current feature distribution, while newly labeled samples in each AL cycle provide fresh supervision that continuously adjusts prototypes to encompass new domain variations, effectively mitigating the distribution gap between source and target domains. To further ensure the distinctiveness and diversity of the learned prototypes, we introduce two auxiliary losses during the iterative optimization.

\subsubsection{Prototype contrastive loss}
To facilitate feature-prototype alignment and discrimination between normal and defective samples, we optimize the spatial relationship between $\mathbf{f}$ and the prototype. Formally, let $y \in \{0,1\}$ denote the label (0 for normal, 1 for defect). The contrastive loss $\mathcal{L}_{cont}$ is defined as:

\begin{equation}
\mathcal{L}_{cont} =
\begin{cases}
\|\mathbf{f} - \mathbf{p}_n^{*}\|_2 + \max\big(0,\, m - \|\mathbf{f} - \mathbf{p}_d^{*}\|_2 \big), & y = 0 \\
\|\mathbf{f} - \mathbf{p}_d^{*}\|_2 + \max\big(0,\, m - \|\mathbf{f} - \mathbf{p}_n^{*}\|_2 \big), & y = 1
\end{cases}
\end{equation}

\noindent where $\mathbf{p}_n^{*}=\arg\min_{\mathbf{p} \in \mathcal{P}_{n}} \|\mathbf{f} - \mathbf{p}\|_2$,  $\mathbf{p}_d^{*}$ is defined analogously, and $m=1$ is the margin boundary. This loss pulls the sample closer to the prototypes of its ground-truth class while pushing it away from the prototypes of the opposite class, thereby enhancing inter-class separability.

\subsubsection{Prototype dispersion loss}
To prevent prototype collapse and promote intra-class diversity, we introduce a prototype dispersion regularization:

\begin{equation}
\mathcal{L}_{disp} = 
\sum_{i < j}^{K_n} \exp\!\left( -\frac{\|\mathbf{p}_n^{(i)} - \mathbf{p}_n^{(j)}\|_2}{\tau} \right) + 
\sum_{i < j}^{K_d} \exp\!\left( -\frac{\|\mathbf{p}_d^{(i)} - \mathbf{p}_d^{(j)}\|_2}{\tau} \right).
\end{equation}

\noindent where $\tau>0$ is a temperature hyperparameter. Minimizing $\mathcal{L}_{disp}$ encourages large pairwise distances within each set of prototypes, thus preventing prototype collapse. The total loss is defined as:

\begin{equation}
\mathcal{L}_{total} = \mathcal{L}_{det} + \lambda_1 \mathcal{L}_{cont} + \lambda_2 \mathcal{L}_{disp},
\end{equation}

\noindent where $\mathcal{L}_{det}$ denotes the detection loss, and the hyperparameters are set to $\lambda_1=1.0$ and $\lambda_2=0.1 $. All parameters, including prototypes and scalars $\alpha, \beta$, are learned end-to-end.

\subsection{Dual-entropy Defect Uncertainty Evaluator}
To enable real-time assessment of adversarial risks, we introduce DeUE to evaluate the uncertainty of each bounding box. As shown in \cref{fig:Image3} (b), for each object query, the detection head outputs the classification probability $p_{c}$, the center coordinates $(x, y)$, and the boundary offset probabilities $p_{l}$ for $(l,t,r,b)$. From $p_{c}$ and $p_{l}$ we derive the confidence $conf$, classification entropy $H_{cls}$ and localization entropy $H_{loc}$ to jointly quantify each box's uncertainty. The formulas are:

\begin{equation}
\begin{cases}
conf = max(p_c) \\
{H}_{cls} = -\sum_{c=1}^{C}p_c \cdot log(p_c) \\
{H}_{loc} = -\frac{1}{4} \sum_{k \in \{l,t,r,b\}}\sum_{l=1}^{K}p_l^k \cdot log(p_{l}^k)
\end{cases}
\end{equation}

\noindent Here $p_{l}^{k}$ denotes the output of the localization head with $K=16$ bins, and $p_c$ denotes the output of the classification head ($C=5$). We normalize both entropies by their theoretical maxima.

\begin{equation}
\tilde{H}_{cls} = \frac{H_{cls}}{log\ C}, \quad 
\tilde{H}_{loc} = \frac{H_{loc}}{log\ K}
\end{equation}

We define the box-level defect uncertainty $U_{d}^{(box)}$ as a weighted aggregation of these normalized components:

\begin{equation}
U_{d}^{(box)} = \frac{1}{w_1+w_2+1} \left(w_1\tilde{H}_{cls}+w_2\tilde{H}_{loc}+1-conf \right)
\end{equation}

For images with multiple detected objects, we define $U_{d}$ as:
\begin{equation}
U_{d} = \max_{b \in \mathcal{B}} \, U_{d}^{(box)}(b)
\end{equation}

\noindent where $\mathcal{B}$ is the set of all predicted boxes. Consequently, $U_{d} \in[0,1]$, with higher values indicating a greater defect uncertainty in detection.

\subsection{Fuzzy Dual-dimension Uncertainty Active Learning}
To balance model reliability and efficiency in dynamic industrial inspection, we employ AL to prioritize the sampling of images that are expected to maximize detector gains, namely false-negative (FN) and false-positive (FP) samples. The correlation between uncertainties ($U_{g}/U_{d}$) and these errors is shown in \cref{fig:Image2} (b). Accordingly, we propose FuDU, which uses a Fuzzy Inference System (FIS) to dynamically assign sampling probabilities.

\subsubsection{Fuzzification and Membership Functions}
As shown in \cref{fig:Image3} (c), FuDU first fuzzifies normalized inputs $U_g$ and $U_d$ (both $\in [0,1]$) using trapezoidal membership functions. Each input is mapped to four linguistic variables: Very Low (VL), Low (L), High (H), and Very High (VH). The membership function $\mu(x)$ for a trapezoidal set defined by parameters $(a, b, c, d)$ is given by:

\begin{equation}
\mu(x; a, b, c, d) = \max \left( 0, \min \left( \frac{x-a}{b-a}, 1, \frac{d-x}{d-c} \right) \right)
\label{eq:important}
\end{equation}

\noindent where the parameters $(a,b,c,d)$ for each linguistic variable are initialized based on nuclear inspection safety standards and fine-tuned on a pilot validation set. The precise parameter values used in this study are listed in the supplementary material.

\subsubsection{Rule Base and Inference}
The output space is defined by four fuzzy sets corresponding to sampling actions: Do Not Sample (DNS), Low-frequency Sample (LS), High-frequency Sample (HS), and Must Sample (MS). The fuzzy rule base follows a "Safety-First" principle designed in collaboration with domain experts. The complete rule matrix is provided in \cref{fig:Image3} (c), governed by three core logic principles:
\begin{itemize}
    \item \textbf{Critical Risk Override:} If either $U_g$ or $U_d$ is VH, the output is MS. This ensures novel defects or severe domain shifts are never missed (100\% sampling).
    \item \textbf{Certainty Suppression:} If both uncertainties are VL or L, the output is DNS or LS. This minimizes annotation costs for confident, normal samples.
    \item \textbf{Balanced Exploration:} If one uncertainty is H while the other is L/VL, the output is HS. This balances cost and potential gain for borderline cases.
\end{itemize}
Following the Mamdani inference method, inputs undergo fuzzification, rule activation (using min-operator), and aggregation (using max-operator) in \cref{eq:important}.

\subsubsection{Defuzzification and Probability Mapping}
To ensure deterministic handling of high-risk samples while maintaining stochasticity for exploration, we employ a hybrid defuzzification strategy. First, we apply the max membership principle to select the optimal output linguistic set $S^* = \arg\max_{S \in \{DNS, LS, HS, MS\}} \mu_{output}(S)$. This avoids the smoothing effect of Centroid defuzzification, ensuring that `Must Sample' cases are not probabilistically diluted. Then, the selected set $S^*$ is mapped to a fixed sampling probability $P_{sample}$: $P_{DNS}=0, P_{LS}=0.1, P_{HS}=0.5, P_{MS}=1.0$. For each image, a random number $r \sim \mathcal{U}(0,1)$ is generated. The image is selected for annotation if $r < P_{sample}$. This procedure yields continuous expected sampling rates while guaranteeing that critical defects (MS) are always captured.

In summary, our FuDU transforms raw detection metrics into interpretable sampling actions through fuzzy logic. By bridging quantitative uncertainty with qualitative expert rules, FuDU establishes a robust mechanism for high-stakes inspection that balances annotation efficiency with reliability.

\section{Experimental evaluation}
\subsection{Dataset and Experimental Setup}
To validate FuDU, we build a nuclear fuel rod defect dataset, split into three subsets: I. 1000 labeled images for initial model training and prototype library construction; II. a 1000-image test set for post-iteration evaluation; III. 2000 unlabeled images, partitioned into four 500-image batches to simulate real-world streaming defect inspection data. The dataset encompasses four categories: abrasion, dent, scratch, and foreign objects, with statistics and details provided in the supplementary material.

In evaluation, the FuDU framework executes a practical workflow: it iteratively conducts sampling and active learning on each batch, with subsequent validation on Subset II. All experiments are conducted using PyTorch on RTX 4090 GPU, with images preprocessed to 640×640 pixels.

\begin{table}[htbp]
  \caption{Detection Results ($mAP_{50}$) of Different AL Methods on the Nuclear Fuel Rod Defect Dataset. The best scores are highlighted in \textbf{bold}.}
  \label{tab:table1}%
  \centering
  \resizebox{\linewidth}{!}{
  \begin{tabular}{p{5.5em}|cccc|ccccccc}
  \toprule
    Method & R1 & R2   & R3    & R4  & Mean  & P (\%)  & R (\%)   & F1  & Ratio & Stream & FPS\\
\midrule
Random  & 83.8$\pm{\scriptstyle 1.1}$ & 84.3$\pm{\scriptstyle 0.9}$ & 84.5$\pm{\scriptstyle 1.1}$ & 85.3$\pm{\scriptstyle 1.0}$ & 84.5  & 15.3 & 16.6 & 15.9 & 16.0\% & $\checkmark$ & 16.8\\
Conf  & 85.1$\pm{\scriptstyle 1.4}$  & 86.8$\pm{\scriptstyle 0.9}$ & 88.0$\pm{\scriptstyle 0.5}$ & 88.7$\pm{\scriptstyle 0.5}$ & 87.2  & 69.3 & 63.4 & 66.2 &13.5\%  & $\checkmark$ & 16.8\\
QBC \cite{QBC} & \textbf{86.7}$\pm{\scriptstyle 2.0}$ & 88.8$\pm{\scriptstyle 1.8}$ & 89.8$\pm{\scriptstyle 1.1}$ & 89.9$\pm{\scriptstyle 0.6}$  & 88.8  & 74.1 & 72.5 & 73.3 & 14.5\%  & \textbf{×} & 0.9\\
SPENet \cite{SPENet} & 86.0$\pm{\scriptstyle 1.8}$ & 87.7$\pm{\scriptstyle 1.4}$ & 90.9$\pm{\scriptstyle 0.8}$ & 93.5$\pm{\scriptstyle 0.4}$  & 89.5  & 49.9 & 87.5 & 63.6 & 25.9\%  & \textbf{×} & 4.4\\
PPAL \cite{PPAL}  & 83.7$\pm{\scriptstyle 0.9}$ & 84.0$\pm{\scriptstyle 1.0}$ & 84.8$\pm{\scriptstyle 0.6}$ & 85.1$\pm{\scriptstyle 0.7}$  & 84.4  & 15.9  &15.3 & 15.6 & 16.0\%  & \textbf{×} & 6.9\\
\textit{Oracle}  & 87.1$\pm{\scriptstyle 1.9}$ & 92.9$\pm{\scriptstyle 1.6}$ & 95.8$\pm{\scriptstyle 0.7}$ & 96.4$\pm{\scriptstyle 0.1}$ & 93.1  & 14.8 & 100.0 & 25.7 & 100.0\% & $\checkmark$ & 16.8\\
\midrule
FuDU  & 86.4$\pm{\scriptstyle 1.4}$   & \textbf{91.9}$\pm{\scriptstyle 1.5}$ & \textbf{94.3}$\pm{\scriptstyle 0.8}$ & \textbf{96.0}$\pm{\scriptstyle 0.3}$ & \textbf{92.2}  & \textbf{95.1} & \textbf{99.3} & \textbf{97.1} & 15.4\% & $\checkmark$ & 12.3\\
\bottomrule
\end{tabular} }
\end{table}%

\subsection{Main Results}
\subsubsection{Quantitative Analysis of Detection Performance}
\cref{tab:table1} compares the performance of FuDU with six other methods.  The evaluation covers standard strategies such as \textbf{Random} sampling (at a fixed ratio) and \textbf{Conf} (using a confidence threshold of $0.2<o<0.8$), alongside the ensemble-based \textbf{QBC}, which selects samples via detector discrepancy across three models. We also include two novel pool-based methods: \textbf{SPENet}, which employs an FN prediction module to estimate box counts, and \textbf{PPAL}, which ranks samples by fusing uncertainty with diversity. The \textbf{Oracle} setting represents the theoretical upper bound, where all images are fully annotated. We set the \textit{random} and score-based \textit{PPAL} ratio at 16\%, corresponding to 320 labeled images (close to the 295 defective samples).

To account for the stochasticity in streaming order and batch partitioning, we repeat each experiment with five random seeds (shuffling the subset III before partitioning into four 500-image batches). We report the mean ± standard deviation of $mAP_{50}$ for each round, the average $mAP_{50}$, the precision of defect detection (P), recall (R), F1 score and the sampling ratio. This protocol ensures that our conclusions are not biased by a specific data order and reflect the method's stability under realistic streaming conditions. The results show that: \textbf{A}. Random sampling fails to support model updates, showing minimal per-round improvement ($<1$mAP) and low precision/recall $<$20\%. This indicates poor selection, with many normal samples mistaken for uncertain cases. \textbf{B}. Confidence-based methods yield better gains, but still miss false negatives. \textbf{C}. Among pool-based AL methods, QBC and SPENet achieve high recall and steady improvements, while PPAL underperforms due to poor scenario adaptability. \textbf{D}. \textit{Oracle} provides the performance upper bound but is not strictly an AL method, as it requires manual review and labeling of all images, leading to low efficiency and high costs. \textbf{E}. Our FuDU achieves the optimal trade-off, enabling significant per-round improvements with a 99.3\% recall rate for defects, making it ideal for high-reliability scenarios such as nuclear fuel rod defect detection. In terms of inference efficiency, FuDU is surpassed only by Random, Conf, and Oracle, which incur no additional inference cost. It outperforms Faster R-CNN-based methods such as PPAL and SPENet, and achieves a 13× speedup over QBC, which requires multi-model ensemble inference.

\begin{figure}[!tb]
\includegraphics[width=\linewidth]{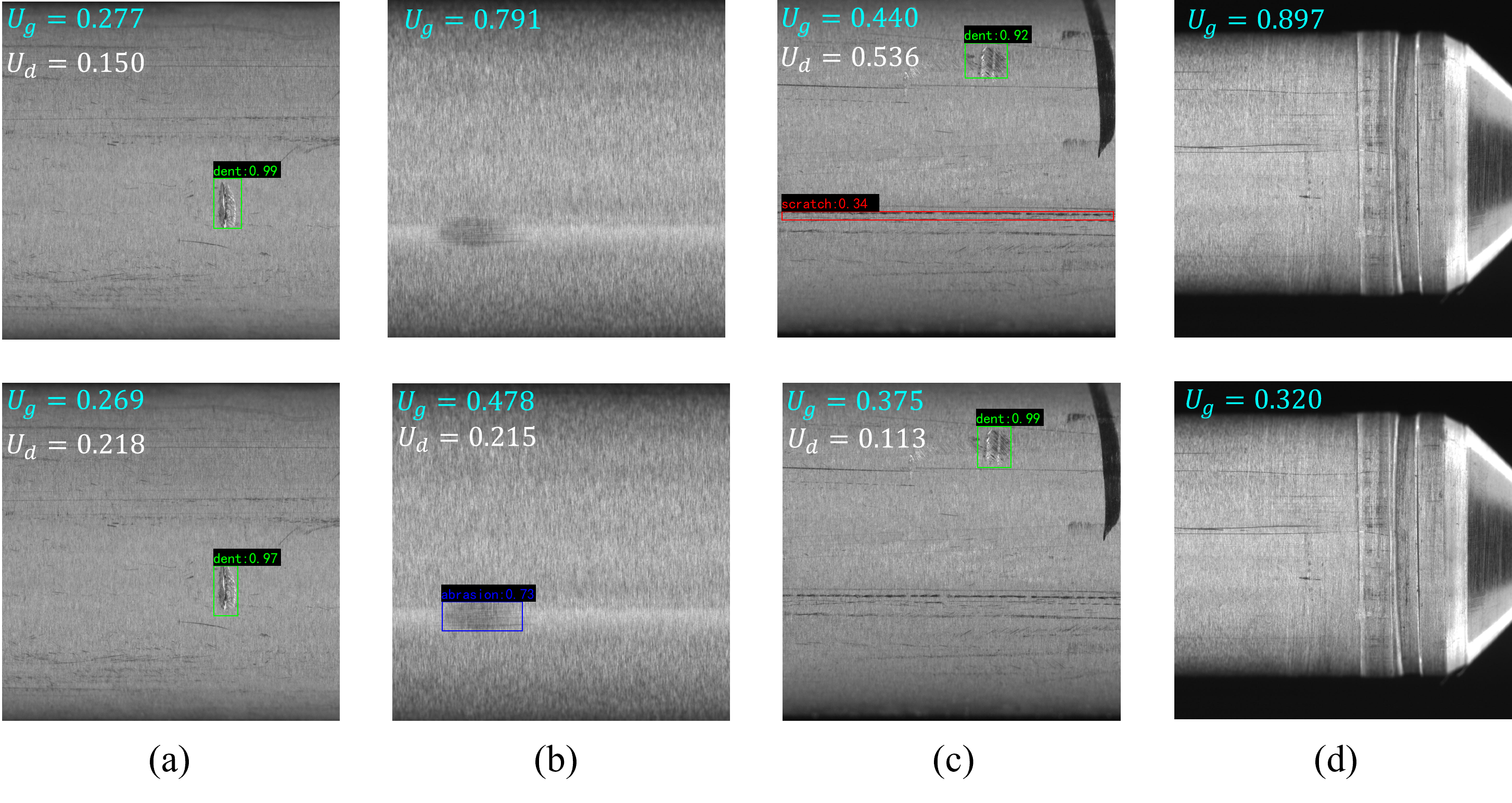}
\caption{Visualization of uncertain samples and detection improvements across FuDU-based AL iterations.}
\label{fig:Image4}
\end{figure}

\subsubsection{Qualitative Analysis of Uncertainty Evolution}
\cref{fig:Image4} visualizes the detection results and uncertainty scores of images before (Row 1) and after (Row 2) active learning. For \cref{fig:Image4} (a), $U_g$ is 0.277, corresponding to Low (L) uncertainty in the fuzzy linguistic variables; $U_d$ is 0.150, lying on the boundary between Very Low (VL) and Low (L) with a membership degree of 0.5 for each. Based on fuzzy inference, the inferred action is Low-frequency Sample (LS), i.e., a 10\% sampling probability. Although this image was not selected, re-inference after active learning (Row 2) shows the model still accurately identifies the defect, with $U_g$ even decreasing. This is because active learning updated the prototypes, making the model more certain about images with such backgrounds. For \cref{fig:Image4} (b), $U_g$ is 0.791, corresponding to Very High (VH) uncertainty. In this case, the model is highly likely to miss detections (False Negatives), so the corresponding fuzzy rule is Must Sample (MS). After active learning, re-inference shows its global uncertainty dropped to 0.478. More importantly, the model can now detect the abrasion defect with lower uncertainty ($U_d=0.215$). \cref{fig:Image4} (c) illustrates another circumstance in which high $U_d$ indicates that the model is uncertain about the detected defects. Clearly, in Row 1, the high $U_d$ corresponds to the red bounding box misclassified as a scratch defect. After active learning, this error is corrected and both $U_g$ and $U_d$ decrease. \cref{fig:Image4} (d) confirms that sampling normal samples with high $U_g$ is also necessary, as prototype updates help reduce the model's uncertainty regarding domain shifts.

\subsection{Ablation Study}
\subsubsection{Global uncertainty} 
To comprehensively evaluate the effectiveness of the PGUQ module, we conduct multi-dimensional ablation studies. First, we investigate the sensitivity to the number of prototype clusters $k$. As shown in Table~\ref{tab:table2}, setting $k$ too small ($k=10$) results in insufficient coverage of the feature space. This causes many samples to be perceived as outliers, leading to inflated uncertainty scores and consequently an excessively high annotation ratio (40.5\%), despite a competitive mean mAP of 92.0\%. Conversely, an excessively large number of prototypes ($k=100$) makes the prototypes difficult to update effectively during training, resulting in a sampling ratio that is too low (9.8\%) to capture informative samples, which degrades the mean mAP to 88.5\%. The optimal balance is achieved at ($k=50$), yielding the highest mean mAP of 92.2\% with an efficient annotation ratio of 15.4\%.

Second, we compare the prototype update strategy: end-to-end learnable static prototypes ($k=50$) significantly outperform re-clustering to generate prototypes after each AL round ($k=50^*$), with mean mAP dropping from 92.2\% to 90.8\% and the annotation ratio increasing from 15.4\% to 21.5\%. This indicates that maintaining prototype stability helps the model better adapt to domain shifts and avoids distribution oscillations caused by frequent reconstruction.

Finally, we compare our prototype-based method against two reconstruction-based uncertainty estimation approaches. A Convolutional Decoder (CD) reconstructs the image $X'$ from features $F$, using the $X-X'$ error as $U_{g}$. A Feature Autoencoder (FAE) reconstructs the feature map $F'$ from $F$, using the $F-F'$ error. The results indicate that both methods exhibit limited efficiency and fail to effectively improve the mean average precision.

\begin{table}[tb]
  \caption{Impact of Prototype Cluster Number $k$, Update Strategy, and Reconstruction-based Baselines. The best mAP scores are highlighted in \textbf{bold}.}
  \label{tab:table2}%
  \centering
  \resizebox{0.5\linewidth}{!}{
  \begin{tabular}{l|cccc|ccc}
  \toprule
    $U_g$ & R1 & R2   & R3    & R4  & Mean   & Ratio & FPS\\
\midrule
$k$=10  & \textbf{86.8}  & \textbf{92.7}  & 93.5 & 94.9 & 92.0  & 40.5\% & 13.1\\

$k$=20  & 86.1 & 91.9 & 92.9 & 93.2 & 91.0  & 31.7\% & 12.9\\

$k$=20*  & 86.1 & 86.8 & 88.0 & 89.7 & 87.7  & 36.5\% & 12.9\\

$k$=50  & 86.4 & 91.9 & \textbf{94.3} & \textbf{96.0} & \textbf{92.2}  & 15.4\% & 12.3\\

$k$=50*  & 86.4 & 91.0 & 92.8 & 93.1  & 90.8  & 21.5\% & 12.3\\

$k$=100 & 84.7 & 88.8 & 90.0 & 90.4  & 88.5  & 9.8\% & 8.8\\
\midrule
CD  & 85.8 & 88.7 & 90.8 & 93.4 & 89.7  & 25.6\% & 7.0\\

FAE  & 86.0  & 87.9 & 89.3 & 91.8 & 88.8   & 20.4\% & 9.3\\
\bottomrule
    \end{tabular}}
\end{table}%

\subsubsection{Defect uncertainty} 
To investigate the contribution of each component in the DeUE, we conduct ablation studies on the weight configuration $(w_1, w_2)$ for classification entropy $\tilde{H}_{cls}$ and localization entropy $\tilde{H}_{loc}$. As shown in Table~\ref{tab:table3}, the first row represents sampling based solely on global uncertainty $U_g$ without evaluating defect uncertainty $U_d$ (Conf=×), yielding a mean mAP of 89.6\%. The second row incorporates confidence scores alongside \(U_g\) (i.e., \(w_1=0, w_2=0\)), slightly improving performance to 90.2\%. However, introducing entropy terms proves more effective. Configurations using confidence combined with localization entropy ($w_1=0$, $w_2=1$) or equal weights for all three terms $w_1=1$, $w_2=1$ achieve competitive results (91.2\% and 91.7\% respectively). Notably, when $w_1$ is excessively large $w_1=2$, $w_2=1$, performance declines to 89.8\%. This is because confidence and classification uncertainty are positively correlated, both reflecting category ambiguity, whereas they form a complementary relationship with localization uncertainty. Over-emphasizing $w_1$ leads to redundant information weighting. Conversely, over-weighting localization entropy ($w_1=1$, $w_2=4$) maintains high accuracy (92.1\%) but significantly increases the annotation ratio to 17.3\%, raising the labeling cost. The configuration $w_1=1$, $w_2=2$ achieves the optimal balance, delivering the highest mean mAP of 92.2\% with an efficient annotation ratio of 15.4\%. Therefore, we adopt $w_1=1$, $w_2=2$ as the default setting.

\begin{table}[tb]
  \caption{Impact of Classification Entropy Weight $w_1$ and Localization Entropy Weight $w_2$. The best mAP scores are highlighted in \textbf{bold}.}
  \label{tab:table3}%
  \centering
  \resizebox{0.5\linewidth}{!}{
  \begin{tabular}{p{3.7em}cc|cccc|ccccc}
  \toprule
    $Conf$ & $w_1$ & $w_2$ & R1 & R2   & R3    & R4  & Mean   & Ratio\\
\midrule
\textbf{×} & 0 & 0  & 85.8 & 88.3 & 91.6 & 92.7      & 89.6   & 14.6\%  \\

$\checkmark$ & 0 & 0   & 86.0 & 90.2 & 91.6 & 93.1      & 90.2  & 15.0\%  \\
$\checkmark$ & 1 & 0   & 85.8 & 90.1 & 91.8 & 93.5      & 90.3  & 14.9\%  \\
$\checkmark$ & 0 & 1  & 86.3 & 90.4 & 93.0 & 94.9      & 91.2   & 15.8\%  \\
$\checkmark$ & 1 & 1  & \textbf{87.0} & 91.3 & 93.2 & 95.4      & 91.7   & 16.0\%  \\
$\checkmark$ & 2 & 1   & 84.5 & 90.0 & 91.5 & 93.3      & 89.8  & 16.5\% \\
$\checkmark$ & 1 & 2   & 86.4 & \textbf{91.9} & \textbf{94.3} & \textbf{96.0}      & \textbf{92.2} & 15.4\%  \\
$\checkmark$ & 1 & 4   & 86.9 & 91.6 & 94.0 & 95.7      & 92.1  & 17.3\% \\
\bottomrule
    \end{tabular}  }
\end{table}%

\begin{table}[tb]
  \centering
  \caption{Ablation Study of Sampling Strategies. Values in parentheses denote mAP improvements over the initial detector, and the best results are highlighted in \textbf{bold}.}
  \label{tab:table4}%
  \resizebox{0.5\linewidth}{!}{
    \begin{tabular}{l|cccccc }
    \toprule
    Sampling Strategies     & R4 ($\uparrow$) & Mean & P(\%) & R(\%) & Ratio \\
    \midrule
    Bilinear        & 92.2 (9.0)   & 89.1  & 42.3  & 79.0  & 27.6\%  \\
    Quadratic       & 93.0 (9.8)   & 89.2   & 63.1   & 84.7   & 19.8\%   \\
    Threshold       & 95.6 (12.4)   & 92.0   & 88.2   & 98.6   & 16.5\%   \\
    Sigmoid         & 94.7 (11.5)   & 90.9   & 61.7   & 97.6   & 23.4\%   \\
    \midrule
    FuDU         & \textbf{96.0 (12.8)}   & \textbf{92.2}  & \textbf{95.1}  & \textbf{99.3}  & \textbf{15.4\%}  \\
    \bottomrule
    \end{tabular}   }
\end{table}%

\subsubsection{Sampling strategies} 
As shown in Table~\ref{tab:table4}, FuDU is compared with four typical sampling strategies under identical settings. It significantly outperforms all baselines, achieving 96.0\% mAP and 99.3\% recall with only 15.4\% annotated samples. This indicates an optimal balance between accuracy, recall, and annotation efficiency, validating its precision in selecting informative samples. The Hard-Threshold method, despite competitive performance, suffers from rigid partitioning and threshold sensitivity, easily missing valuable boundary samples. In contrast, FuDU enables finer-grained sample ranking via fuzzy rules. Among smoothing functions, Quadratic and Sigmoid outperform Bilinear, confirming that suppressing low-uncertainty responses is crucial for sampling quality. Notably, Sigmoid achieves 97.6\% recall (close to Hard-Threshold) but requires more annotations (23.4\% ratio) than FuDU. In summary, FuDU achieves state-of-the-art detection performance at low annotation costs, proving the effectiveness and superiority of its designed uncertainty fusion mechanism. The values in parentheses following R4 denote the performance improvement (mAP) relative to the model's initial evaluation on the held-out test set before any active learning iterations.

\subsection{Generalization and Flexibility Experiments}
\subsubsection{Flexibility Validation with Different Batch Settings}
To verify the flexibility and practicality of the proposed FuDU framework, we further adopt a finer-grained setting for the original 2000 images: 10 batches with 200 images per batch. The experimental results are shown in \cref{tab:table5}, which demonstrates that the FuDU framework maintains stable and consistent detection performance under both batch configurations. This indicates that the proposed method can flexibly adapt to various data partitioning strategies without performance degradation. Such flexibility is beneficial for practical industrial deployment, since the scale and batch arrangement of input data often vary in real-world scenarios.

\begin{table}[tb]
  \caption{Detection Results ($mAP_{50}$) under the Finer-grained 10-Batch Streaming Setting. The best AL mAP scores are highlighted in \textbf{bold}, while \textit{Oracle} denotes the full-annotation upper bound.}
  \label{tab:table5}%
  \centering
  \resizebox{\linewidth}{!}{
  \begin{tabular}{p{5.5em}|cccccccccc|c}
  \toprule
    Method & R1 & R2 & R3 & R4 & R5 & R6 & R7 & R8 & R9 & R10 & Ratio\\
\midrule
Random        & 83.3$\pm{\scriptstyle 0.1}$ & 83.6$\pm{\scriptstyle 0.2}$ & 83.8$\pm{\scriptstyle 0.3}$ & 84.1$\pm{\scriptstyle 0.5}$ & 84.4$\pm{\scriptstyle 0.7}$ & 84.5$\pm{\scriptstyle 0.8}$ & 84.9$\pm{\scriptstyle 0.9}$ & 85.0$\pm{\scriptstyle 1.1}$ & 85.1$\pm{\scriptstyle 1.3}$ & 85.4$\pm{\scriptstyle 1.1}$ & 16.0\% \\

Conf          & 83.9$\pm{\scriptstyle 0.2}$ & 84.4$\pm{\scriptstyle 0.3}$ & 85.0$\pm{\scriptstyle 0.4}$ & 85.6$\pm{\scriptstyle 0.3}$ & 86.7$\pm{\scriptstyle 0.1}$ & 87.8$\pm{\scriptstyle 0.9}$ & 88.3$\pm{\scriptstyle 0.8}$ & 88.8$\pm{\scriptstyle 0.7}$ & 89.0$\pm{\scriptstyle 0.4}$ & 88.9$\pm{\scriptstyle 0.2}$ & 10.9\% \\

QBC \cite{QBC}  & 83.4$\pm{\scriptstyle 0.2}$ & 83.8$\pm{\scriptstyle 0.4}$ & 84.4$\pm{\scriptstyle 0.5}$ & 85.6$\pm{\scriptstyle 0.5}$ & 88.1$\pm{\scriptstyle 1.1}$ & 87.8$\pm{\scriptstyle 0.8}$ & 89.5$\pm{\scriptstyle 0.4}$ & 89.7$\pm{\scriptstyle 0.4}$ & 89.8$\pm{\scriptstyle 0.3}$ & 90.1$\pm{\scriptstyle 0.2}$ & 16.3\% \\

SPENet \cite{SPENet} & \textbf{84.0}$\pm{\scriptstyle 0.3}$ & 84.9$\pm{\scriptstyle 0.2}$ & 85.5$\pm{\scriptstyle 0.6}$ & 87.0$\pm{\scriptstyle 1.1}$ & 87.9$\pm{\scriptstyle 0.4}$ & 90.4$\pm{\scriptstyle 0.4}$ & 91.9$\pm{\scriptstyle 0.4}$ & 92.1$\pm{\scriptstyle 0.5}$ & 92.9$\pm{\scriptstyle 0.3}$ & 93.7$\pm{\scriptstyle 0.2}$ & 28.8\% \\

PPAL \cite{PPAL} & 83.7$\pm{\scriptstyle 0.5}$ & 83.6$\pm{\scriptstyle 0.2}$ & 83.9$\pm{\scriptstyle 0.5}$ & 84.5$\pm{\scriptstyle 0.4}$ & 84.5$\pm{\scriptstyle 0.4}$ & 84.5$\pm{\scriptstyle 0.4}$ & 84.8$\pm{\scriptstyle 0.3}$ & 84.9$\pm{\scriptstyle 0.3}$ & 85.3$\pm{\scriptstyle 0.3}$ & 85.3$\pm{\scriptstyle 0.2}$ & 16.0\% \\

\textit{Oracle} & 85.1$\pm{\scriptstyle 0.4}$ & 86.4$\pm{\scriptstyle 0.6}$ & 87.5$\pm{\scriptstyle 0.6}$ & 88.8$\pm{\scriptstyle 0.8}$ & 89.9$\pm{\scriptstyle 1.4}$ & 91.8$\pm{\scriptstyle 1.0}$ & 93.1$\pm{\scriptstyle 0.3}$ & 95.0$\pm{\scriptstyle 0.3}$ & 96.1$\pm{\scriptstyle 0.2}$ & 96.6$\pm{\scriptstyle 0.1}$ & 100.0\% \\

\midrule
FuDU            & \textbf{84.0}$\pm{\scriptstyle 0.4}$ & \textbf{85.6}$\pm{\scriptstyle 1.0}$ & \textbf{86.8}$\pm{\scriptstyle 1.4}$ & \textbf{88.8}$\pm{\scriptstyle 1.4}$ & \textbf{90.8}$\pm{\scriptstyle 1.1}$ & \textbf{91.5}$\pm{\scriptstyle 0.9}$ & \textbf{92.3}$\pm{\scriptstyle 1.1}$ & \textbf{94.3}$\pm{\scriptstyle 1.0}$ & \textbf{95.5}$\pm{\scriptstyle 0.2}$ & \textbf{96.1}$\pm{\scriptstyle 0.1}$ & 14.9\% \\
\bottomrule
    \end{tabular} }
\end{table}

\begin{table}[tb]
  \centering
  \caption{Cross-Architecture Generalization of FuDU-Based and Confidence-Based AL Strategies. Values in parentheses denote mAP improvements over the initial detector, and the better results within each architecture are highlighted in \textbf{bold}.}
  \resizebox{0.6\linewidth}{!}{
  \begin{tabular}{llc|ccccc}
    \toprule
    Backbone & Head & Method & R4 ($\uparrow$) & Mean & P(\%) & R(\%) & Ratio \\
    \midrule
    \multirow{2}{*}{Swin \cite{Swin}} & RT-DETR & Conf  & 90.0 (4.8) & 87.3 & 59.1 & 60.3 & 15.1\% \\
                          &     & FuDU  & \textbf{92.5 (7.3)} & \textbf{91.6} & \textbf{93.7} & \textbf{97.3} & 15.0\% \\
    \midrule
    \multirow{2}{*}{ViTDet \cite{VITDET}} & RT-DETR & Conf  & 86.3 (4.2) & 84.2 & 63.6 & 65.8 & 15.3\% \\
                           &  & FuDU  & \textbf{91.1 (9.0)} & \textbf{90.3} & \textbf{93.2} & \textbf{98.3} & 15.6\% \\
    \midrule
    \multirow{2}{*}{RN50} & DEIM \cite{DEIM} & Conf  & 85.7 (5.0) & 83.4 & 66.9 & 67.8 & 15.0\% \\
                            &  & FuDU  & \textbf{89.9 (9.2)} & \textbf{87.6} & \textbf{93.0} &\textbf{ 99.0} & 15.7\% \\
    \midrule
    \multirow{2}{*}{RN50} & MR-DETR \cite{MRDETR}& Conf  & 88.9 (4.3) & 87.8 & 69.1 & 63.7 & 13.6\% \\
                             &  & FuDU  & \textbf{94.3 (9.7)} & \textbf{92.4} & \textbf{92.4} & \textbf{99.3} & 15.9\% \\
    \midrule
    \multirow{2}{*}{RN50} & MI-DETR \cite{MIDETR} & Conf  & 88.5 (3.9) & 86.6 & 83.3 & 64.9 & 11.1\% \\
                                &  & FuDU  & \textbf{95.6 (11.2)} & \textbf{93.2} & \textbf{95.4} & \textbf{99.0} & 15.3\% \\

    \bottomrule
    \end{tabular}  }
  \label{tab:table6}%
\end{table}

\subsubsection{Modularity and Generality Across Different Backbones and Detectors}
FuDU computes uncertainty at both the backbone and detection head, making it compatible with other detectors of similar architectures. To validate its generalizability, we extend FuDU to Transformer based backbones (Swin \cite{Swin}, ViTDet \cite{VITDET}) and SOTA detection heads (DEIM \cite{DEIM}, MR-DETR \cite{MRDETR}, MI-DETR \cite{MIDETR}) comparing its performance against a confidence-based AL baseline. The results in Table~\ref{tab:table6} confirm that FuDU enables efficient model iteration across diverse detectors, demonstrating clear cross-architecture generalization.

\subsubsection{Cross-Dataset Generalization}
We further validate FuDU on the public ELES dataset \cite{SSN} via four AL rounds using 2,000 photovoltaic cell inspection images, including 800 defective samples. For cross-dataset transfer, FuDU keeps the fuzzy rule base and sampling-probability mapping unchanged, while only lightly recalibrating the membership boundaries according to the target uncertainty distribution. The results in Table~\ref{tab:table7} show that FuDU achieves the most substantial performance gains among all AL methods, with a recall rate ($>$97\%) second only to $Oracle$. This indicates that FuDU can generalize to different industrial inspection scenarios without exhaustive expert rule redesign. Detailed configurations are provided in the supplementary material.

\begin{table}[tb]
  \caption{Cross-Dataset Generalization Results ($mAP_{50}$) on the Public ELES Dataset. The best AL results are highlighted in \textbf{bold}, while \textit{Oracle} denotes the full-annotation upper bound.}
  \label{tab:table7}%
  \centering
  \resizebox{0.8\linewidth}{!}{
  \begin{tabular}{p{5.3em}|cccc|cccc}
  \toprule
    Method & R1 & R2   & R3  & R4  & Mean  & P (\%)  & R (\%)   & Ratio \\
\midrule
Random  & 75.8$\pm{\scriptstyle 0.9}$ & 78.0$\pm{\scriptstyle 1.0}$ & 80.4$\pm{\scriptstyle 0.8}$ & 83.3$\pm{\scriptstyle 2.4}$  & 79.4  & 40.5 & 40.5  & 40\% \\
Conf  & 76.6$\pm{\scriptstyle 1.6}$  & 79.7$\pm{\scriptstyle 0.7}$ & 82.0$\pm{\scriptstyle 0.9}$ & 86.4$\pm{\scriptstyle 1.0}$   & 81.2 & 65.4 & 48.6  & 29.8\% \\
QBC \cite{QBC} & 77.7$\pm{\scriptstyle 2.5}$ & 81.9$\pm{\scriptstyle 1.1}$ & 85.5$\pm{\scriptstyle 1.4}$ & 89.0$\pm{\scriptstyle 1.2}$  & 83.5 & \textbf{90.0} & 72.1  & 32.1\% \\
SPENet \cite{SPENet} & 77.2$\pm{\scriptstyle 1.0}$ & 82.1$\pm{\scriptstyle 1.6}$ & 85.0$\pm{\scriptstyle 1.4}$ & 88.6$\pm{\scriptstyle 0.9}$  & 83.2 & 69.5 & 59.2  & 47.0\% \\
PPAL \cite{PPAL}  & 77.8$\pm{\scriptstyle 1.8}$ & 80.8$\pm{\scriptstyle 1.2}$ & 82.6$\pm{\scriptstyle 1.1}$ & 85.7$\pm{\scriptstyle 1.5}$  & 81.7 & 41.8 & 41.8  & 40\% \\
\textit{Oracle}  & 80.9$\pm{\scriptstyle 0.6}$ & 84.5$\pm{\scriptstyle 0.5}$ & 89.7$\pm{\scriptstyle 0.8}$ & 91.2$\pm{\scriptstyle 0.2}$   & 86.6 & 40.0 & 100.0 & 100\% \\
\midrule
FuDU  & \textbf{80.5}$\pm{\scriptstyle 1.4}$  & \textbf{84.9}$\pm{\scriptstyle 0.8}$ & \textbf{87.0}$\pm{\scriptstyle 0.3}$ & \textbf{90.4}$\pm{\scriptstyle 0.1}$   & \textbf{ 85.7} & 86.4 & \textbf{97.1}  & 45.0\% \\
\bottomrule
    \end{tabular} }
\end{table}%

\section{Conclusion}
This paper addresses the challenge of reliable defect detection with limited annotations by proposing FuDU, a stream-based active learning method built on a fuzzy dual-dimensional uncertainty framework. FuDU features a Prototype-based Global Uncertainty Quantification (PGUQ) module and a Dual-Entropy Uncertainty Evaluator (DeUE), which integrate data uncertainty with domain knowledge to form an expert-informed sampling strategy. Experiments show that FuDU achieves state-of-the-art performance—96.0\% mAP and 99.3\% recall—while requiring only 15.4\% of samples, significantly surpassing existing strategies. In conclusion, FuDU meets the stringent needs for real-time, reliable defect detection and offers a practical pathway for intelligent quality inspection systems to evolve from passive updating to active self-evolution.

\section*{Acknowledgements}
This work was supported in part by the National Natural Science Foundation of China No. 62473127, the S\&T Program of Hebei Province No. 242Q4302Z, the National Key Research and Development Program of China No. 2024YFB3310904, the Interdisciplinary Postgraduate Training Program of Hebei University of Technology No.HEBUT-Y-XKJC-2023002, and the China Scholarship Council No. 202506700027.

%
%
\bibliographystyle{splncs04}
\bibliography{main}
\end{document}